\documentclass[runningheads]{llncs}
\usepackage[T1]{fontenc}
\usepackage[misc]{ifsym}
\usepackage{graphicx}
\usepackage{multirow}
\usepackage{amsmath,amssymb,amsfonts}
\usepackage{color}
\usepackage[colorlinks,bookmarksopen,bookmarksnumbered,citecolor=blue, linkcolor=red, urlcolor=blue]{hyperref}
\begin{document}
	\title{Robust Semi-supervised Multimodal \\ Medical Image Segmentation via \\ Cross Modality Collaboration}
	\titlerunning{Multimodal Medical Image Segmentation via Cross Modality Collaboration}
	%
	
\author{Xiaogen Zhou\inst{1} \and
Yiyou Sun \inst{2} \and
Min Deng \inst{2} \and
\\ Winnie Chiu Wing Chu \inst{2}$^{(\textrm{\Letter})}$ \and
Qi Dou \inst{1}$^{(\textrm{\Letter})}$ }
	%
	\authorrunning{X. Zhou et al.}
	%
	\institute{\textsuperscript{1}Department of Computer Science and Engineering, \\ The Chinese University of Hong Kong, Hong Kong, China\\ \textsuperscript{2} Department of Imaging and Interventional Radiology, \\ The Chinese University of Hong Kong, Hong Kong, China}
	\maketitle              
	\begin{abstract}

Multimodal learning leverages complementary information derived from different modalities, thereby enhancing performance in medical image segmentation. However, prevailing multimodal learning methods heavily rely on extensive well-annotated data from various modalities to achieve accurate segmentation performance. This dependence often poses a challenge in clinical settings due to limited availability of such data. Moreover, the inherent anatomical misalignment between different imaging modalities further complicates the endeavor to enhance segmentation performance. To address this problem, we propose a novel semi-supervised multimodal segmentation framework that is robust to scarce labeled data and misaligned modalities. Our framework employs a novel cross modality collaboration strategy to distill modality-independent knowledge, which is inherently associated with each modality, and integrates this information into a unified fusion layer for feature amalgamation. With a channel-wise semantic consistency loss, our framework ensures alignment of modality-independent information from a feature-wise perspective across modalities, thereby fortifying it against misalignments in multimodal scenarios. Furthermore, our framework effectively integrates contrastive consistent learning to regulate anatomical structures, facilitating anatomical-wise prediction alignment on unlabeled data in semi-supervised segmentation tasks. Our method achieves competitive performance compared to other multimodal methods across three tasks: cardiac, abdominal multi-organ, and thyroid-associated orbitopathy segmentations. It also demonstrates outstanding robustness in scenarios involving scarce labeled data and misaligned modalities. Code is available at: \href{https://github.com/med-air/CMC}{https://github.com/med-air/CMC}.

\end{abstract}
	
\section{Introduction}
	
Multimodal learning is a long-standing topic with great importance for medical image analysis. It leverages synergistic information of multiple modalities to outperform single-modality solutions. Many multimodal methods have been developed for medical image segmentation. These methods either concatenate a large array of well-annotated multiple Magnetic Resonance Imaging (MRI) and computed tomography (CT) as input \cite{b7_1,b2_3}, or fuse higher-level information from pre-aligned modalities within a latent space \cite{b3_3,b0_2}. However, in real-world clinical settings, the availability of extensive collections of meticulously annotated and anatomically aligned datasets from various modalities cannot be always guaranteed. This is primarily due to the expensive costs associated with detailed annotation and a shortage of experienced clinicians available for labeling tasks. Consequently, robustness to sparsely labeled data or misaligned modalities during inference is essential for a widely-applicable multimodal learning method.

A typical solution to address the scarcity of labeled data is to employ semi-supervised learning techniques \cite{b1_2,b1_1}. These methods can effectively leverage the limited labeled data in conjunction with abundant unlabeled data to train deep learning models, thereby mitigating the dependence on costly labeled data while maintaining satisfactory performance. However, most existing semi-supervised learning methods are inherently constrained to single-modality input, failing to capitalize on the wealth of multimodal data frequently encountered in daily clinical routines. Furthermore, Zhu et al. \cite{b5_1} and Zhang et al. \cite{b7_2} have attempted to address this by applying semi-supervised learning to minimize consistency loss on abundant unlabeled data across different modalities. However, different modalities vary in intensity distributions and appearances, and simply minimizing consistency loss may not be sufficient for effective feature fusion and alignment. To address this challenge, a recent work \cite{b3_3} introduces a modality-collaborative semi-supervised framework. By leveraging modality-specific knowledge, this method endeavors to enhance multimodal image segmentation, thereby facilitating more effective feature fusion and alignment.

For the effective extraction of modality-invariant representations that capture critical information and enable feature fusion across various modalities, it is beneficial to identify and leverage modality-specific knowledge. This can be accomplished through modality-independent feature extraction, which distills the characteristics and semantic representations of each modality \cite{b3_3,b3_5}. Additionally, integrating a modality-collaborative strategy can significantly enhance feature fusion and alignment across different modalities, ensuring effective synergistic integration of multimodal data \cite{b3_3,b7_1}. However, these works fail to enforce constraints to regularize the underlying anatomical structures during the feature fusion stage, which is crucial for maintaining alignment. 
\begin{figure}[h]
\centering
{\includegraphics[width=12.00cm,height=3.9375cm]{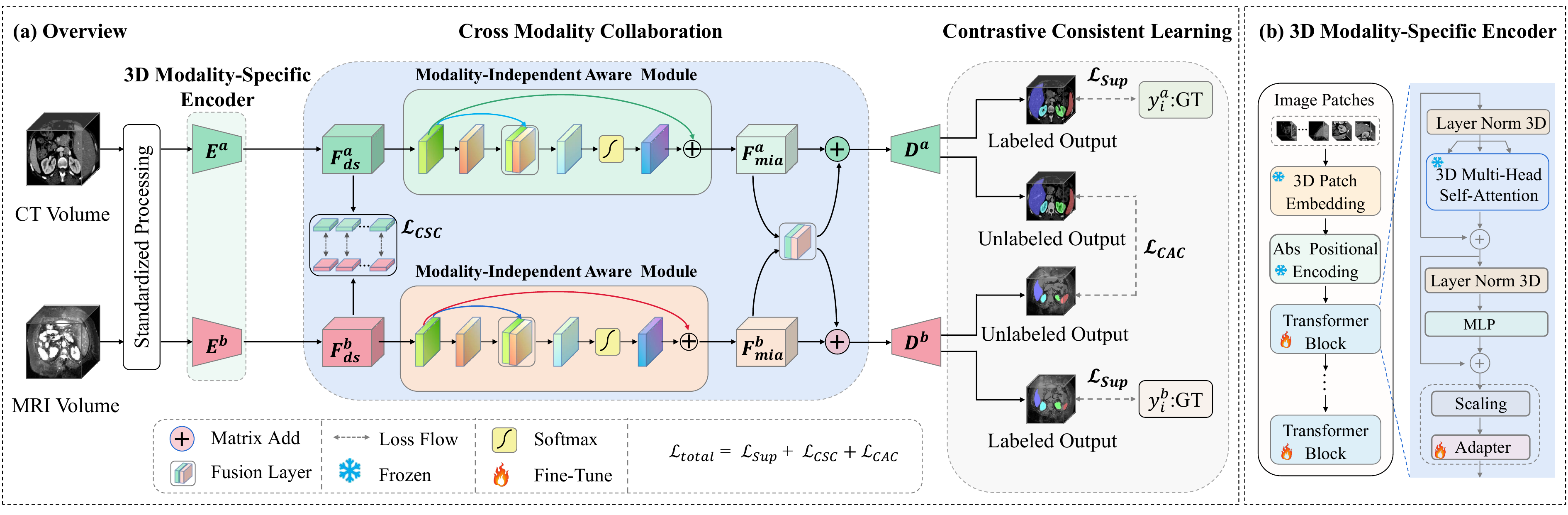}}
\vspace{.00cm}
\caption{ The pipeline of our framework. It consists of three main components. (1) Two 3D foundation model-driven encoders are employed to capture 3D volumetric representations from different modalities. (2) A novel cross-modality collaborative strategy is introduced to fuse information and align feature maps. (3) A contrastive consistent learning module is exploited to generate consistent predictions for unlabeled data.}
\label{fig_1_flowchart}
\end{figure}

In this paper, we propose a novel semi-supervised multimodal framework that incorporates cross modality collaboration and contrastive consistent learning. This framework is robust to scenarios with limited labeled data and misaligned modalities. Our framework exploits a novel cross modality-collaboration strategy to regularize the underlying channel-wise features and distill modality-independent knowledge. The modality-independent knowledge extracted from each modality is fused to a common fusion layer, which contains collaborative information for segmentation. To effectively extract modality-independent information, we design modality-specific encoders to capture the distinctive semantic features intrinsic to each modality. Moreover, we introduce a novel contrastive consistent learning module to mitigate inconsistencies and align anatomical structures between prediction maps from unlabeled data across different modalities, thereby facilitating contrastive consistent learning process among image modalities. We validate our proposed method on the task of 3D semi-supervised multimodal medical image segmentation across three datasets for different tasks. The results demonstrate that our method achieves competitive performance compared to state-of-the-art methods for misaligned modality data and exhibits high robustness in situations with scarce labeled data.

\section{Methodology}
An overview of our framework is presented in Fig. \ref{fig_1_flowchart}. We first employ two 3D modality-specific encoders to get the initial modality-independent features from multimodal inputs. Then, we introduce a cross-modality collaboration strategy to facilitate feature fusion and channel-wise alignment. Subsequently, we introduce a contrastive consistent learning module aimed at reducing inconsistencies in anatomical structures across prediction maps derived from the unlabeled data. The details of our method are elaborated in the following sections.
\subsection{Cross Modality Collaboration for Alignment and Feature Fusion}
We denote the multimodal images by $\mathcal{D}^{l} = \{\{(x_{l,i}^{a}, y_{i}^{a})\}_{i=1}^N, \{(x_{l,i}^{b}, y_{i}^{b})\}_{i=1}^N\}$ and $\mathcal{D}^{u}=\{\{(x_{u,j}^{a})\}_{j=1}^M, \{(x_{u,j}^{b})\}_{j=1}^M\}$, where $x_{l,i}$ and $x_{u,j}$ denote the labeled and unlabeled data in modality $\text{a}$ and $\text{b}$, respectively. The $y_{i}^a$ and $y_{i}^b$ are the corresponding segmentation masks for $x_{l,i}^a$ and $x_{l,i}^b$. The $N$ and $M$ denote the number of labeled and unlabled samples, and $N \ll M$. Each modality $x_i^a$ and $x_i^b$ is input to modality-specific encoder $E^a$ and $E^b$, respectively, which are fine-tuned from the pre-trained encoder of SAM-Med3D \cite{b8_0}. Moreover, as shown in Fig. \ref{fig_1_flowchart} (b), we integrate an adapter module within each Vision Transformer (ViT) block. This adapter serves as a bottleneck infrastructure and consists of a sequence of two 3D convolutional layers with different kernel sizes. 

Then, we obtain its distinctive semantic (DS) features by $F_{ds}^a=E^a(x_i^a)$ and $F_{ds}^b=E^b(x_i^b)$.	
For effective extraction of distinctive semantic features $F_{ds}^a$ and $F_{ds}^b$, which are expected to regularize underlying anatomical structures from a perspective of channel-wise feature alignment across modalities, we introduce a Channel-wise Semantic Consistency (CSC) loss which is used to align the channel-wise features. In specific, the $\mathcal{L}_{\text{CSC}}$ loss can be defined as follows:
	\begin{equation}\label{CSC_loss_eq1}
         \small
	\mathcal{\mathcal{L}}_{\text{CSC}} = \min \limits_{E^a, E^b} ~ \sum_{c}-\log\frac{\exp(Sim((F_{ds}^a)_c, (F_{ds}^b)_c))}{\sum_{c} \exp(Sim((F_{ds}^a)_c, (F_{ds}^b)_c))}  ,
	\end{equation}
where $c\in \{1,2,...,C\}$ denotes the $c_{th}$ channel of two features, i.e., $F_{ds}^a$ and $F_{ds}^b$, respectively. $Sim(.,.)$ represents the cosine similarity \cite{b8_3} used to measure the channel-wise semantic feature similarity between the latent representations.

After the feature-wise alignment procedure, which mitigates the influence of modality-specific appearance features from multimodal data, we introduce a novel Modality-Independent Awareness (MIA) module to further harness modality independent knowledge from each modality for effective feature fusion. As is known from literature of deep neural networks, modality-independent knowledge is useful in learning generalized and robust representations across various imaging modalities \cite{b3_3,b3_6}. As depicted in Fig. \ref{fig_1_flowchart}, the MIA module integrates modality-aware attention mechanisms \cite{b2_3}, 3D convolutional layers, and a fusion layer. Subsequently, the features $F_{mia}^a$ and $F_{mia}^b$ generated by each MIA module are subsequently merged in a fusion layer to produce the fused feature.
\subsection{Contrastive Consistent Learning with Contrastive Consistency}
Contrastive consistency has shown effectiveness in semi-supervised learning \cite{b3_3,b7_2}. The formulation of the contrastive consistency term is based on the correlation and differential information across various modalities. The consistency of their predictions acts as the impetus for semi-supervised cross-modal knowledge reciprocal learning. This also holds true for our scenario. We first build two decoders $D^a$ and $D^b$, which are fine-tuned the pre-trained foundation model decoder \cite{b4_1} on our datasets. During training, the weights of our model are optimized using a supervised loss function $\mathcal{\mathcal{L}}_{\text{Sup}}$ on labeled data, which is defined as follows:
	\begin{equation} \label{ses_loss}
        \small
	\mathcal{\mathcal{L}}_{\text{Sup}}=\min\limits_{E^{a}, E^{b}, D^{a}, D^{b}} ~ \mathbb{E}_{x_{l,i}^{a},x_{l,i}^{b}}[\mathcal{\mathcal{L}}_{\text{CE}}(D^a(E^a(x_{l,i}^{a})),y_i^a) + \mathcal{\mathcal{L}}_{\text{Dice}}(D^b(E^b((x_{l,i}^{b}),y_i^b)))],
	\end{equation}
where $\mathcal{\mathcal{L}}_{\text{CE}}$ and $\mathcal{\mathcal{L}}_{\text{Dice}}$ represent cross-entropy loss function and Dice coefficient loss function. To further regularize underlying anatomical structures from a perspective of anatomical-wise predictions alignment on unlabeled data across modalities for multimodal segmentation, we introduce a Contrastive Anatomical-similar Consistency (CAC) loss. This loss is designed to measure the anatomical similarity between predictions of different modalities on unlabeled data, thereby improving the accuracy and robustness of multimodal segmentation. The CAC loss incorporates the Dice coefficient similarity $Sim_{Dice}$ \cite{b7_2} as a metric for assessing anatomical similarity. The $\mathcal{\mathcal{L}}_{\text{CAC}}$ loss is defined as follows: 
	\begin{equation}\label{cac_loss_eq1}
        \small
	 \mathcal{\mathcal{L}}_{\text{CAC}} =\min\limits_{E^a, E^b,D^{a}, D^{b}}  \sum_{j} -\log\frac{\exp(Sim_{Dice}(D^a(E^a(x_{u,j}^{a})), D^b(E^b(x_{u,j}^{b})))}{\sum_{j=1}^\text{M} \exp(Sim_{Dice}(D^a(E^a(x_{u,j}^{a})), D^b(E^b(x_{u,j}^{b})))}.
	\end{equation}
Overall, the total loss of our framework is $\mathcal{L}_{\text{total}}=\mathcal{L}_{\text{Sup}} + \alpha \mathcal{L}_{\text{CSC}} + \beta \mathcal{L}_{\text{CAC}}$.
To balance between these losses, we adopt a ramp-up weighting coefficient $\alpha = 0.1 \cdot e^{-5(1-\frac{t}{t_{\text{max}}})}$ and a ramp-down weighting coefficient $\beta = 0.1 \cdot e^{-5(\frac{t}{t_{\text{max}}})}$. This strategy follows related works \cite{b5_2,b5_1}, where $t$ and $t_{\text{max}}$ denote the current and maximum number of epochs, respectively. All approaches were implemented using PyTorch on multiple NVIDIA GPUs. We trained the models using the Adam optimizer with an initial learning rate of $10^{-5}$. The input size is 96 $\times$ 96 $\times$ 96 voxels and batch size is 4.

\section{Experiments}

\subsection{Datasets} 

We extensively evaluated our framework on three datasets of multi-modal image segmentation, including two public challenge datasets and one private dataset. 

\textbf{MS-CMRSeg Dataset.} The MS-CMRSeg \cite{b6_1} dataset consists of cardiac magnetic resonance images from 45 patients with cardiomyopathy. The dataset includes three modalities: late gadolinium enhancement (LGE), T2-weighted (T2w) and balanced-steady state free precession (BSSFP). Notably, the dataset are paired as they originate from the same patients and are registered. The ground truth of the left ventricular cavity (LV), right ventricular cavity (RV), and left ventricular myocardium (Myo) is provided. For our experiment, we used 40 LGE/BSSFP image pairs for training and 5 pairs for testing and validation.

\textbf{AMOS Dataset.} The AMOS dataset \cite{b6_2} comes from a multi-modality abdominal multi-organ segmentation challenge. It includes 300 CT and 60 MRI images from multi-center, multi-modality and multi-phase, each annotations includes 15 abdominal organs. Notably, the CT/MRI data of AMOS dataset is unpaired as they originate from different patients. The training set consists of 200 CT and 40 MRI images, and the testing set comprises 100 CT and 20 MRI images. In our experiments, we randomly selected 40 CT images from the 200 available in the training set and included 40 MRI images, creating a new training set. Additionally, we randomly selected 20 CT images from the 100 available in the testing set and included 20 MRI images, forming a new test and validation set.

\textbf{TAO Dataset.} The thyroid-associated orbitopathy (TAO) dataset is an in-house multimodal dataset of thyroid-associated orbitopathy collected from the Gerald Choa Neuroscience Centre MRI Core Facility at the Prince of Wales Hospital in Hong Kong. This dataset comprises 100 cases, each of which underwent orbital MRI and received a definitive TAO diagnosis. Each case includes pre-contrast T1-weighted (T1) and fat-suppressed post-contrast T1-weighted (T1c), which are sequentially acquired from the same patient. Notably, the T1/T1c data in the TAO dataset is unpaired due to differences in in-plane resolution. All MRI was performed on a 3.0 T  Siemens scanner with a Head/Neck 64 Channel coil. T1 data was acquired using volumetric interpolated breath-hold examination (VIBE) pulse sequence with in-plane resolution of 0.555 $\times$ 0.555 mm\(^2\) and slice thickness of 1.5 mm. T1c data was collected using a fat-suppressed spoiled-gradient echo (GRE) core sequence with in-plane resolution of 0.625 $\! \times \!$ 0.625 mm\(^2\) and slice thickness of 1.5 mm. The manual annotation was performed by a trained rater using ITK-SNAP, under the guidance of a senior radiologist who has over 20 years of experience.  The segmentation mask includes 8 anatomical structures of extraocular muscles, i.e., herniation of the lacrimal gland (LG), and compression and edema of the optic nerve (ON), inferior oblique muscle (IOM), superior oblique muscle (SOM), superior rectus (SR), lateral rectus (LR), medial rectus (MR), and inferior rectus (IR). For our experiments, we randomly selected 80 T1/T1c image pairs from 100 cases as the training set, and the remaining 20 T1/T1c image pairs for the testing and validation set.

	\begin{table}[h]\tiny
		\caption{ Quantitative results of our approach and other methods on the AMOS \cite{b6_2} and TAO datasets in terms of Dice score with 10\% labeled data.}
		\label{AMOS_TAO_seg_result}
		\centering
\begin{tabular}{llccccccccc}
\hline
\multicolumn{11}{c}{\textbf{AMOS 2022 dataset}}                                                                                                                                                                                                                                                                                                                        \\ \hline
\multicolumn{1}{l|}{\multirow{2}{*}{\textbf{Methods}}} & \multicolumn{5}{c|}{\textbf{CT}}                                                                                                                                          & \multicolumn{5}{c}{\textbf{\textbf{MRI}}}                                                                                                                     \\ \cline{2-11} 
\multicolumn{1}{l|}{}                        & \multicolumn{1}{c}{\textbf{Spleen}}            & \textbf{R.Kidney}           & \textbf{L.Kidney}           & \multicolumn{1}{c|}{\textbf{Liver}}              & \multicolumn{1}{c|}{\textbf{Avg.}}    & \textbf{Spleen}              & \textbf{R.Kidney}          & \textbf{L.Kidney}                             & \multicolumn{1}{c|}{\textbf{Liver}}              & \textbf{Avg.}   \\ \hline
\multicolumn{1}{l|}{\textbf{Sup}\cite{b7_0}}                     & \multicolumn{1}{c}{61.2$\pm$0.6}          & 64.1$\pm$1.2          & 66.9$\pm$1.1           & \multicolumn{1}{c|}{65.7$\pm$1.5}         & \multicolumn{1}{c|}{65.7}          & 68.1$\pm$1.2          & 76.2$\pm$0.8         & 67.8$\pm$1.2                              & \multicolumn{1}{c|}{71.5$\pm$1.2}         & 70.4          \\
\multicolumn{1}{l|}{\textbf{EFCD}\cite{b2_2}}                    & \multicolumn{1}{c}{69.4$\pm$0.8}          & 65.3$\pm$1.1           & 69.7$\pm$0.4          & \multicolumn{1}{c|}{73.5$\pm$1.3}         & \multicolumn{1}{c|}{70.4}          & 71.2$\pm$0.4          & 78.2$\pm$1.1          & 70.3$\pm$1.4                              & \multicolumn{1}{c|}{74.7$\pm$0.7}          & 73.8          \\
\multicolumn{1}{l|}{\textbf{UMML}\cite{b5_1}}                    & \multicolumn{1}{c}{71.4$\pm$1.2}          & 66.3$\pm$1.2          & 70.2$\pm$1.3           & \multicolumn{1}{c|}{75.3$\pm$0.7}          & \multicolumn{1}{c|}{71.5}          & 75.8$\pm$1.7           & 76.7$\pm$0.8          & 72.1$\pm$1.9                              & \multicolumn{1}{c|}{76.2$\pm$1.5}          & 76.4          \\
\multicolumn{1}{l|}{\textbf{mmFor}\cite{b7_1}}                   & \multicolumn{1}{c}{72.6$\pm$1.3}          & 73.5$\pm$0.5          & 73.1$\pm$1.4           & \multicolumn{1}{c|}{79.7$\pm$1.7}          & \multicolumn{1}{c|}{74.8}          & 76.3$\pm$1.4           & 78.9$\pm$1.2         & 72.2$\pm$1.5                              & \multicolumn{1}{c|}{76.4$\pm$1.8}          & 78.6          \\
\multicolumn{1}{l|}{\textbf{CML}\cite{b7_2}}                     & \multicolumn{1}{c}{73.6$\pm$0.8}          & 75.4$\pm$1.3          & 76.2$\pm$1.7          & \multicolumn{1}{c|}{\textbf{81.2$\pm$1.1}} & \multicolumn{1}{c|}{75.4}          & 77.8$\pm$1.6          & \textbf{81.1$\pm$0.7} & 73.9$\pm$1.7                              & \multicolumn{1}{c|}{77.8$\pm$1.4}          & 82.9          \\ \hline
\multicolumn{1}{l|}{\textbf{Ours}}                   & \multicolumn{1}{c}{\textbf{74.8$\pm$0.9}} & \textbf{76.5$\pm$1.3} & \textbf{77.9$\pm$1.8} & \multicolumn{1}{c|}{80.7$\pm$1.8}         & \multicolumn{1}{c|}{\textbf{76.3}} & \textbf{78.4$\pm$1.8} & 79.5$\pm$1.5          & \textbf{74.7$\pm$1.8}                     & \multicolumn{1}{c|}{\textbf{79.6$\pm$1.3}} & \textbf{84.3} \\ \hline
\multicolumn{11}{c}{\textbf{TAO 2022 dataset}}                                                                                                                                                                                                                                                                                                                         \\ \hline
\multicolumn{1}{l|}{\textbf{Modality}}                & \multicolumn{1}{l|}{\textbf{Methods}}           & \textbf{SR }                 & \textbf{LG }                 & \textbf{IOM}                                     & \textbf{LR}                                 & \textbf{IR}                  & \textbf{MR}                 & \multicolumn{1}{c|}{\textbf{SOM}}               & \multicolumn{2}{c}{\textbf{Avg.}}                                \\ \hline
\multicolumn{1}{l|}{\multirow{6}{*}{\textbf{T1}}}    & \multicolumn{1}{l|}{\textbf{Sup}\cite{b7_0}}              & 71.5$\pm$2.6            & 73.8$\pm$1.3            & 70.6$\pm$3.6                                & 71.3$\pm$3.5                           & 72.2$\pm$3.2            & 66.1$\pm$5.6           & \multicolumn{1}{c|}{45.9$\pm$1.8}          & \multicolumn{2}{c}{64.5}                                \\
\multicolumn{1}{l|}{}                        & \multicolumn{1}{l|}{\textbf{EFCD}\cite{b2_2}}             & 73.1$\pm$8.8            & 75.6$\pm$1.8            & 71.2$\pm$0.7                                & 72.8$\pm$1.5                           & 74.3$\pm$2.7            & 68.8$\pm$4.7           & \multicolumn{1}{c|}{51.7$\pm$9.1}          & \multicolumn{2}{c}{70.4}                                \\
\multicolumn{1}{l|}{}                        & \multicolumn{1}{l|}{\textbf{UMML}\cite{b5_1}}             & 73.4$\pm$3.2            & 77.3$\pm$3.2            & 73.7$\pm$1.3                                & 74.5$\pm$1.1                           & 75.2$\pm$4.2            & 69.1$\pm$0.4           & \multicolumn{1}{c|}{52.3$\pm$1.4}          & \multicolumn{2}{c}{71.6}                                \\
\multicolumn{1}{l|}{}                        & \multicolumn{1}{l|}{\textbf{mmFor}\cite{b7_1}}            & 75.6$\pm$4.3            & 78.7$\pm$2.3            & 75.3$\pm$2.2                                & 75.2$\pm$2.4                           & 77.2$\pm$5.7            & 69.8$\pm$4.2           & \multicolumn{1}{c|}{57.4$\pm$4.3}          & \multicolumn{2}{c}{73.1}                                \\
\multicolumn{1}{l|}{}                        & \multicolumn{1}{l|}{\textbf{CML}\cite{b7_2}}              & 76.7$\pm$3.1           & 80.6$\pm$5.4            & \textbf{79.7$\pm$3.3}                       & 78.8$\pm$2.3                           & 78.3$\pm$3.3            & 71.5$\pm$6.2           & \multicolumn{1}{c|}{62.4$\pm$5.8}          & \multicolumn{2}{c}{75.5}                                \\ \cline{2-11} 
\multicolumn{1}{l|}{}                        & \multicolumn{1}{l|}{\textbf{Ours}}             & \textbf{77.8$\pm$2.3}   & \textbf{81.8$\pm$1.7}   & 78.4$\pm$2.3                                & \textbf{80.2$\pm$3.1}                  & \textbf{81.2$\pm$5.1}   & \textbf{73.4$\pm$1.8}  & \multicolumn{1}{c|}{\textbf{65.6$\pm$2.1}} & \multicolumn{2}{c}{\textbf{76.9}}                                \\ \hline
\multicolumn{1}{l|}{\multirow{6}{*}{T1c}}    & \multicolumn{1}{l|}{\textbf{Sup}\cite{b7_0}}              & 65.5$\pm$3.6            & 67.3$\pm$4.2            & 66.6$\pm$2.2                                & 68.3$\pm$2.5                           & 69.7$\pm$3.2            & 62.1$\pm$4.4           & \multicolumn{1}{c|}{47.5$\pm$4.5}          & \multicolumn{2}{c}{64.3}                                \\
\multicolumn{1}{l|}{}                        & \multicolumn{1}{l|}{\textbf{EFCD}\cite{b2_2}}             & 68.1$\pm$8.8            & 70.2$\pm$3.3            & 70.2$\pm$3.2                                & 71.2$\pm$1.5                           & 72.3$\pm$2.7            & 64.8$\pm$7.7           & \multicolumn{1}{c|}{52.7$\pm$7.1}          & \multicolumn{2}{c}{67.1}                                \\
\multicolumn{1}{l|}{}                        & \multicolumn{1}{l|}{\textbf{UMML}\cite{b5_1}}             & 70.4$\pm$2.4            & 72.3$\pm$2.4            & 72.7$\pm$0.5                                & 73.5$\pm$1.1                           & 74.2$\pm$4.2            & 67.5$\pm$2.4           & \multicolumn{1}{c|}{55.7$\pm$5.4}          & \multicolumn{2}{c}{69.6}                                \\
\multicolumn{1}{l|}{}                        & \multicolumn{1}{l|}{\textbf{mmFor}\cite{b7_1}}            & 73.6$\pm$3.5            & 73.7$\pm$0.5            & 73.3$\pm$0.2                                & 75.3$\pm$0.4                           & 76.2$\pm$5.7            & 70.4$\pm$4.2           & \multicolumn{1}{c|}{59.4$\pm$5.3}          & \multicolumn{2}{c}{71.9}                                \\
\multicolumn{1}{l|}{}                        & \multicolumn{1}{l|}{\textbf{CML}\cite{b7_2}}              & 74.7$\pm$1.3           & 75.2$\pm$5.4            & 75.7$\pm$6.7                                & 77.8$\pm$2.3                           & \textbf{79.3$\pm$3.8}   & 71.2$\pm$6.2           & \multicolumn{1}{c|}{64.1$\pm$0.8}          & \multicolumn{2}{c}{74.1}                                \\ \cline{2-11} 
\multicolumn{1}{l|}{}                        & \multicolumn{1}{l|}{\textbf{\textbf{Ours}}}             & \textbf{75.9$\pm$4.3}   & \textbf{77.8$\pm$3.3}   & \textbf{77.4$\pm$8.1}                       & \textbf{79.2$\pm$5.3}                  & 79.1$\pm$5.1            & \textbf{72.4$\pm$4.7}  & \multicolumn{1}{c|}{\textbf{66.3$\pm$5.4}} & \multicolumn{2}{c}{\textbf{75.7}}                                \\ \hline
\end{tabular}
\end{table}

\begin{figure}[t]
		\centering
		{\includegraphics[width=12cm,height=7.0cm]{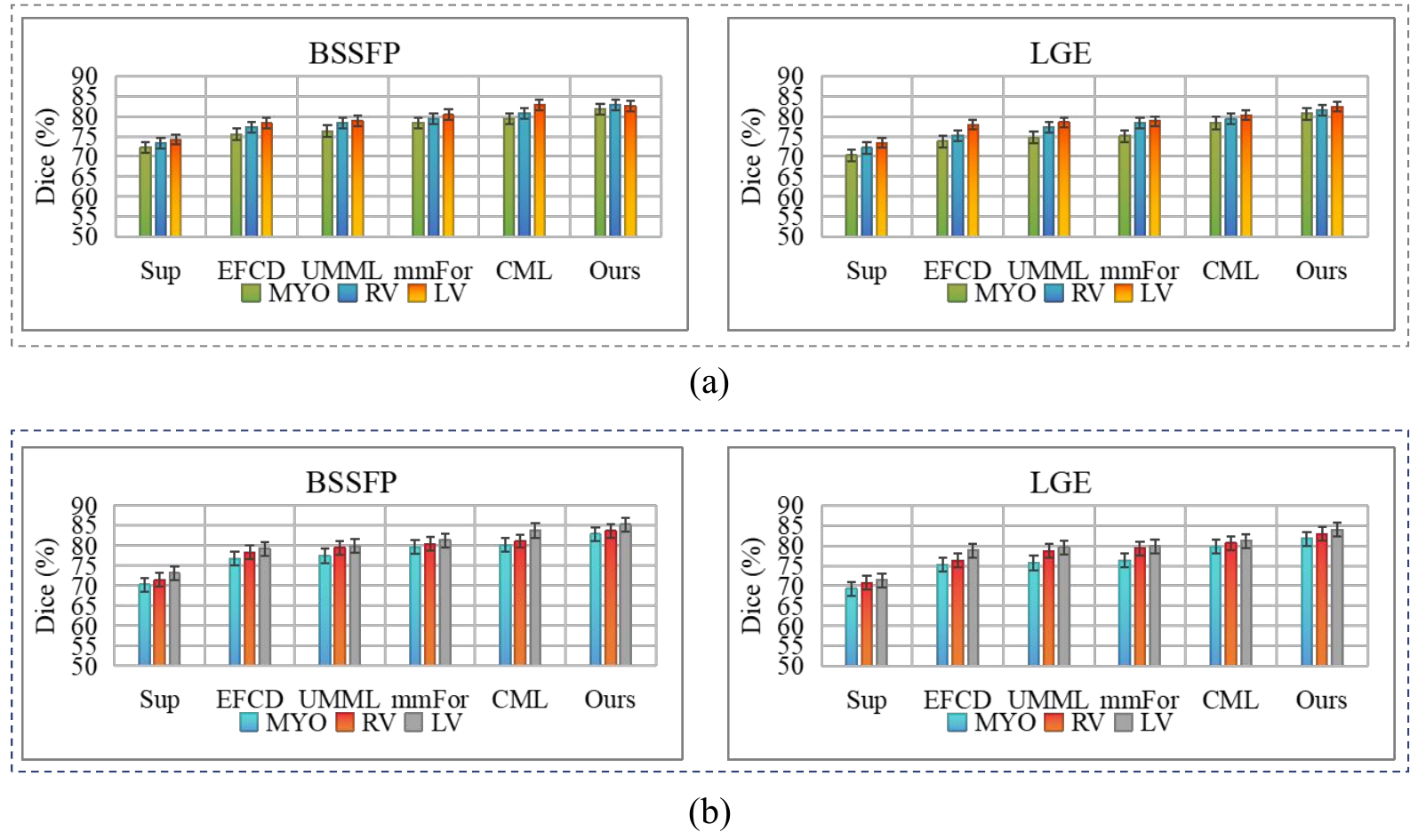}}
		\vspace{.00cm}
		\caption{Comparison of segmentation performance across different models on the MS-CMRSeg \cite{b6_1} dataset using various labeled data ratios, evaluated in terms of Dice score.}
		\label{cardiac_bar_chart}
	\end{figure}

\subsection{Comparison with State-of-the-art Methods}
We used the CLIP-Driven Universal Model pre-trained weights \cite{b4_1} as our backbone when using a single RTX 3090 GPU. When using multiple GPUs, we used the SAM-Med3D \cite{b8_0} pre-trained weights \cite{b4_1} as our backbone. We conducted extensive experiments, using 10\% and 20\% labeled data ratios from three datasets for training, and employed the Dice score and Average Symmetric Surface Distance (ASSD) for quantitative evaluation. We compared our model with other multimodal learning methods, such as EFCD \cite{b2_2} and mmFor \cite{b7_1}. Additionally, we compared our model with semi-supervised multimodal learning approaches, such as UMML \cite{b5_1} and CML \cite{b7_2}, and the fully supervised method V-Net \cite{b7_0}. Table \ref{AMOS_TAO_seg_result} presents the quantitative performance of different methods on the AMOS \cite{b6_2} and TAO datasets. The results show that our framework considerably surpasses the comparison methods in both CT and MRI modalities, achieving high Dice scores in a label-scarce scenario on the AMOS \cite{b6_2}. As is well known, unpaired data from different modalities originate from different patients and cannot be directly aligned. This complicates the extraction of consistent features from different modalities compared to paired data. Despite the AMOS dataset \cite{b6_2} is unpaired, our model demonstrates superior performance. This can be attributed to its architecture, which is based on the Transformer model of SAM-Med3D \cite{b8_0}. The input images are divided into patches, which are then linearly embedded and combined with positional encoding, providing strong feature learning capability. We then utilize the CSC loss to align the channel-wise features from multimodal images. Furthermore, we introduce a novel MIA module to effectively harness modality-independent knowledge from each modality, facilitating efficient feature fusion. Consequently, our model achieves superior performance in semi-supervised multimodal segmentation tasks. For the TAO dataset, specifically for the T1 and T1c modalities, our framework has also shown promising results. Fig. \ref{cardiac_bar_chart} provides a comparative analysis of segmentation performance across various models on the MS-CMRSeg dataset. The performance results evaluated using the ASSD score on the MS-CMRSeg dataset, are presented in Table 3 of the supplementary material. It is evident from the results that our method surpasses the comparison methods in both BSSFP and LGE modalities, consistently achieving high Dice scores.\\
	\begin{figure}[h]
		\centering
		{\includegraphics[width=12cm,height=3.11cm]{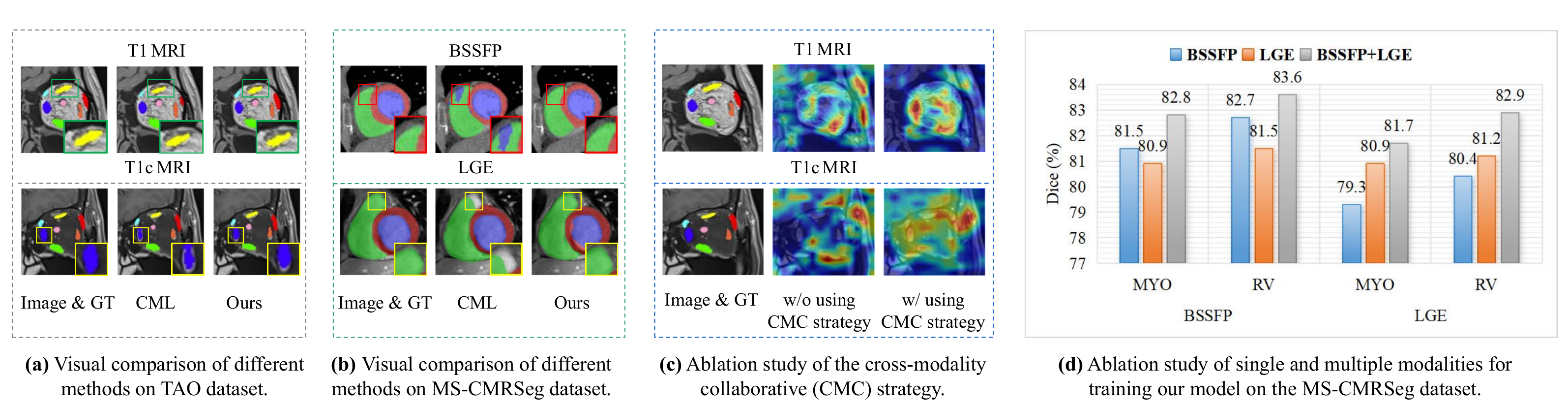}}	\vspace{.00cm}
		\caption{ (a) and (b) Visual comparison between the CML \cite{b7_2} and our method on the TAO and MS-CMRSeg \cite{b6_1} datasets. (c) Ablation analysis of the cross CMC strategy. (d) Ablation study of single and multiple modalities for training our model on the MS-CMRSeg dataset \cite{b6_1}. }
		\label{TAO_cardiac_result_fig}
	\end{figure}

Fig. \ref{TAO_cardiac_result_fig} (a) and (b) present a qualitative comparison of our framework with the CML \cite{b7_2} method for thyroid-associated orbitopathy and cardiac segmentation. The results demonstrate that our model's predictions align more closely with the ground truth and contain fewer erroneous regions compared to CML \cite{b7_2}.
	
	\begin{table}[h]\scriptsize
		\caption{ Ablation results of our method on the AMOS dataset \cite{b6_2}.}
		\label{ablation_seg_result}
		\centering
		\begin{tabular}{cccc|cccc|cccc}
			\hline
			\multicolumn{4}{c|}{\multirow{2}{*}{Methods}}                                                                                                                                                                                                                                                                                      & \multicolumn{4}{c|}{Dice (\%)}                                                                                                                                     & \multicolumn{4}{c}{ASSD (mm)}                                                                                                                                      \\ \cline{5-12} 
			\multicolumn{4}{c|}{}                                                                                                                                                                                                                                                                                                              & \multicolumn{2}{c|}{CT}                                                                    & \multicolumn{2}{c|}{MRI}                                               & \multicolumn{2}{c|}{CT}                                                                    & \multicolumn{2}{c}{MRI}                                                \\ \hline
			\multicolumn{1}{c|}{\multirow{2}{*}{Baseline}} & \multicolumn{1}{c|}{\multirow{2}{*}{\begin{tabular}[c]{@{}c@{}}Modality-Specific\\ Encoder\end{tabular}}} & \multicolumn{1}{c|}{\multirow{2}{*}{\begin{tabular}[c]{@{}c@{}}CMC \\ Stratey\end{tabular}}} & \multirow{2}{*}{\begin{tabular}[c]{@{}c@{}}CCL \\ Module\end{tabular}} & \multicolumn{1}{c|}{\multirow{2}{*}{Spleen}} & \multicolumn{1}{c|}{\multirow{2}{*}{Liver}} & \multicolumn{1}{c|}{\multirow{2}{*}{Spleen}} & \multirow{2}{*}{Liver} & \multicolumn{1}{c|}{\multirow{2}{*}{Spleen}} & \multicolumn{1}{c|}{\multirow{2}{*}{Liver}} & \multicolumn{1}{c|}{\multirow{2}{*}{Spleen}} & \multirow{2}{*}{Liver} \\
			\multicolumn{1}{c|}{}                          & \multicolumn{1}{c|}{}                                                                                     & \multicolumn{1}{c|}{}                                                                        &                                                                        & \multicolumn{1}{c|}{}                        & \multicolumn{1}{c|}{}                       & \multicolumn{1}{c|}{}                        &                        & \multicolumn{1}{c|}{}                        & \multicolumn{1}{c|}{}                       & \multicolumn{1}{c|}{}                        &                        \\ \hline
			\multicolumn{1}{c|}{$\checkmark$}                         & \multicolumn{1}{c|}{}                                                                                     & \multicolumn{1}{c|}{}                                                                        &                                                                        & \multicolumn{1}{c|}{70.4}                    & \multicolumn{1}{c|}{76.7}                   & \multicolumn{1}{c|}{73.3}                    & 74.4                   & \multicolumn{1}{c|}{4.41}                    & \multicolumn{1}{c|}{4.12}                   & \multicolumn{1}{c|}{4.38}                    & 4.16                   \\ \hline
			\multicolumn{1}{c|}{$\checkmark$}                         & \multicolumn{1}{c|}{$\checkmark$}                                                                                    & \multicolumn{1}{c|}{}                                                                        &                                                                        & \multicolumn{1}{c|}{72.9}                    & \multicolumn{1}{c|}{78.5}                   & \multicolumn{1}{c|}{75.9}                    & 76.2                   & \multicolumn{1}{c|}{4.17}                    & \multicolumn{1}{c|}{3.85}                   & \multicolumn{1}{c|}{3.87}                    & 3.73                   \\ \hline
			\multicolumn{1}{c|}{$\checkmark$}                         & \multicolumn{1}{c|}{$\checkmark$}                                                                                    & \multicolumn{1}{c|}{$\checkmark$}                                                                       &                                                                        & \multicolumn{1}{c|}{73.4}                    & \multicolumn{1}{c|}{79.2}                   & \multicolumn{1}{c|}{76.2}                    & 77.3                   & \multicolumn{1}{c|}{3.58}                    & \multicolumn{1}{c|}{3.47}                   & \multicolumn{1}{c|}{3.49}                    & 3.38                   \\ \hline
			\multicolumn{1}{c|}{$\checkmark$}                         & \multicolumn{1}{c|}{$\checkmark$}                                                                                    & \multicolumn{1}{c|}{$\checkmark$}                                                                       &$\checkmark$                                                                     & \multicolumn{1}{c|}{\textbf{74.8}}           & \multicolumn{1}{c|}{\textbf{80.7}}          & \multicolumn{1}{c|}{\textbf{78.4}}           & \textbf{79.6}          & \multicolumn{1}{c|}{\textbf{3.42}}           & \multicolumn{1}{c|}{\textbf{3.38}}          & \multicolumn{1}{c|}{\textbf{3.23}}           & \textbf{3.17}          \\ \hline
		\end{tabular}
	\end{table}
\subsection{Ablation Study}
\label{sec:Ablation}
We investigate the effectiveness of three key components in our method: modality-specific encoder, cross modality collaboration (CMC) strategy, and contrastive consistent learning (CCL) module. As shown in Table \ref{ablation_seg_result}, we first set up a baseline network without using these components. Subsequently, we add the modality-specific encoder, CMC strategy, and CCL module one by one into the baseline network. Integrating these components consistently enhances performance, thereby highlighting their critical significance within our framework. Fig. \ref{TAO_cardiac_result_fig} (c) presents an ablation analysis of our framework on the TAO dataset, comparing the performance with and without using the CMC strategy. Additionally, we conduct an ablation study on the MS-CMRSeg \cite{b6_1} dataset to compare the performance of using single and multiple modalities for training our model, as shown in Fig. \ref{TAO_cardiac_result_fig} (d). The results demonstrate that using multiple modalities (i.e., BSSFP$\pm$LGE) outperforms using a single modality (i.e., BSSFP or LGE). The integration of the CMC strategy into our framework significantly enhances the highlighting of object regions and improves the capacity for class-discriminative localization  compared to frameworks without the CMC strategy. Consequently, this demonstrates that the CMC strategy is able to foster a more efficient feature fusion.
\section{Conclusion}
We propose a novel semi-supervised multimodal segmentation framework that synergistically incorporates cross-modality collaboration and contrastive consistent learning strategies to achieve modality-independent knowledge and feature alignment across different modalities. We validate our method on three benchmark datasets and achieve comparable results to the state-of-the-art approaches. Notably, the framework exhibits exceptional robustness against significant inference variations, underscoring its potential for widespread clinical application in real-world clinical settings. More efficient fusion strategies on 3D medical image datasets and cross-modality settings will be conducted in our future studies.\\

\hspace{-2em} \textbf{Acknowledgements.} This work was supported by Hong Kong Research Grants Council under Projects No. T45-401/22-N and Project No. 14200721.\\

\hspace{-2em} \textbf{Disclosure of Interests.}
The authors have no competing interests to declare that are relevant to the content of this article.

\bibliographystyle{splncs04}
\bibliography{paper_2001}

\end{document}